\documentclass[letterpaper, 10 pt, conference]{ieeeconf}  
\usepackage{amsmath}
\usepackage{graphicx} 
\usepackage[linesnumbered,ruled,vlined]{algorithm2e}
\usepackage{booktabs}  
\usepackage{adjustbox}  
\usepackage{float}  
\usepackage{multirow}
\usepackage{amssymb}
\usepackage{fancyhdr}
\usepackage{url}

\IEEEoverridecommandlockouts                              

\title{\LARGE \bf
SOD-YOLOv8 - Enhancing YOLOv8 for Small Object Detection in Traffic Scenes
}
\author{Boshra Khalili$^{1}$ and Andrew W.Smyth$^{2}$
\thanks{$^{1}$Boshra Khalili is Graduate Research Assistant in the Department of Civil Engineering and Engineering Mechanics, Columbia University, New York, NY 10027, USA  {\tt\small bk2898@columbia.edu}}%
\thanks{$^{2}$Andrew W. Smyth (corresponding author) is the Robert A. W. and Christine S. Carleton Professor of Civil Engineering and Engineering Mechanics and the Director of the Center for the Smart Streetscapes (CS3), Columbia University, New York, NY 10027, USA {\tt\small  aws16@columbia.edu}}%
}

\begin{document}

\maketitle
\thispagestyle{empty}
\pagestyle{empty}
\pagenumbering{roman}
\newpage
\begin{abstract}

 Object detection as part of computer vision can be crucial for traffic management, emergency response, autonomous vehicles, and smart cities. Despite significant advances in object detection, detecting small objects in images captured by distant cameras remains challenging due to their size, distance from the camera, varied shapes, and cluttered backgrounds. To address these challenges, we propose Small Object Detection YOLOv8 (SOD-YOLOv8), a novel model specifically designed for scenarios involving numerous small objects. Inspired by Efficient Generalized Feature Pyramid Networks (GFPN), we enhance multi-path fusion within YOLOv8 to integrate features across different levels, preserving details from shallower layers and improving small object detection accuracy. Additionally, a fourth detection layer is introduced to utilize high-resolution spatial information effectively. The Efficient Multi-Scale Attention Module (EMA) in the C2f-EMA module enhances feature extraction by redistributing weights and prioritizing relevant features. We introduce Powerful-IoU (PIoU) as a replacement for CIoU, focusing on moderate-quality anchor boxes and adding a penalty based on differences between predicted and ground truth bounding box corners. This approach simplifies calculations, speeds up convergence, and enhances detection accuracy. SOD-YOLOv8 significantly improves small object detection, surpassing widely used models in various metrics, without substantially increasing computational cost or latency compared to YOLOv8s. Specifically, it increases recall from 40.1\% to 43.9\%, precision from 51.2\% to 53.9\%, $\text{mAP}_{0.5}$ from 40.6\% to 45.1\%, and $\text{mAP}_{0.5:0.95}$ from 24\% to 26.6\%.  In dynamic real-world traffic scenes, SOD-YOLOv8 demonstrated notable improvements in diverse conditions, proving its reliability and effectiveness in detecting small objects even in challenging environments.

\end{abstract}

\section{INTRODUCTION}

Object detection in computer vision plays a crucial role across various fields, including Autonomous Vehicles \cite{chen2017multiview,alqarqaz2023object,lim2023object}, traffic scene monitoring \cite{feng2023,chuai2023improved}, enhancing intelligent driving systems \cite{salimans2016improved}, and facilitating search and rescue missions \cite{alsamhi2022}. Accurate detection of small objects such as pedestrians, vehicles, motorcycles, bicycles, traffic signs, and lights is crucial for safe navigation and decision-making in autonomous vehicles and intelligent driving systems \cite{chen2017multiview,lim2023object}. Furthermore, detecting small objects enhances traffic flow management, pedestrian safety, and overall traffic scene analysis. This capability is essential for improving urban planning and transportation systems \cite{feng2023,chuai2023improved}.

As the cost of UAV production decreases and flight control techniques advance, these small, flexible devices are increasingly used for intelligent traffic monitoring \cite{wang2023uav}. UAVs typically operate at higher altitudes to capture broader views, which reduces the apparent size of ground objects due to greater distances. This distance complicates object detection within captured images \cite{wang2023uav}. Despite significant progress in object detection, detecting small objects such as pedestrians, motorcycles, bicycles, and vehicles in urban traffic remains challenging due to their size, varied shapes, and cluttered backgrounds. This challenge is further amplified when working with limited hardware resources in computer vision and object detection.

Small objects, which occupy a small portion of an image and have lower resolution and less distinct visual characteristics compared to larger objects, are more challenging to detect accurately. Moreover, shallow layers in networks such as YOLOv8 may filter out essential spatial details required for detecting these small objects, resulting in data loss. Additionally, smaller objects can be overshadowed by larger ones during feature extraction, potentially causing the loss of relevant details crucial for accurate detection. Overcoming these challenges is crucial for improving overall detection accuracy and reliability in real-world scenarios.

To address small object detection in UAV aerial photography and traffic scenes, we introduce a novel model based on YOLOv8. Our model integrates multi-scale spatial and contextual information using an enhanced GFPN \cite{jiang2022giraffedet}.We integrate EMA Attention \cite{ouyang2023efficient} into the C2f module to ensure that small object features are given sufficient emphasis. We also include a fourth detection layer to utilize high-resolution spatial details effectively. Furthermore, due to the significant impact of bounding box regression in object detection, we used the PIoU method, which enhances performance and reduces convergence time by incorporating an improved penalty term and attention mechanism. Our key contributions are as follows:

\begin{itemize}
    \item Inspired by the Efficient RepGFPN in DAMO-YOLO models \cite{xu2022damoyolo}, we enhance multi-path fusion within the YOLOv8 architecture. This enhancement facilitates better fusion of features across different levels and simplifies the GFPN structure through reparameterization. By preserving crucial information from shallower layers, our approach significantly improves detection accuracy, especially for small objects. Additionally, we add a fourth detection layer to effectively leverage high-resolution and detailed spatial information.
    
    \item We integrate a C2f-EMA structure into the network, leveraging the Efficient Multi-Scale Attention Module to replace C2f in the neck layers. This enhancement improves feature extraction by redistributing feature weights, prioritizing relevant features and spatial details across image channels. Consequently, it enhances the network's ability to detect targets of different sizes. 
    
    \item We use PIoU to replace CIoU in the original network. PIoU enhances existing IoU-based loss functions by better balancing difficult and easy samples, with a specific focus on anchor boxes of moderate quality. It incorporates a penalty term based on differences between predicted and ground truth bounding box corners, improving the efficiency and accuracy of bounding box regression. PIoU also simplifies calculations by requiring only one hyperparameter, leading to faster convergence and enhanced object detection accuracy.
    
   \item We conduct visual analyses on various challenging scenarios to demonstrate the effectiveness of our proposed approach in enhancing small object detection. Additionally, we perform experiments in real-world traffic scenes using images captured from cameras mounted on buildings. These images contain numerous small objects and help validate our enhanced model for small object detection.
\end{itemize}

The structure of the remaining sections of this paper is as follows: Section 2 discusses related work. Section 3 provides an overview of the YOLOv8 network architecture. Section 4 details the proposed enhanced YOLOv8. Section 5 covers our experimental setup and result analysis. Finally, Section 6 concludes the paper.

\section{Related work}
Small object detection has been a significant challenge in the field of computer vision, particularly in traffic scenarios. This section reviews mainstream object detection algorithms, recent advancements in small object detection, and specific enhancements made to the YOLO framework.

Mainstream object detection algorithms predominantly use deep learning techniques, categorized into two types: two-stage and one-stage methods. Two-stage methods process candidate frames with a classifier and perform deep learning on corresponding frames \cite{girshick2014rich}. Typical two-stage detection algorithms include R-CNN \cite{girshick2014rich}, Fast R-CNN \cite{girshick2015fastrcnn}, and Faster R-CNN \cite{ren2015fasterrcnn}. The R-CNN family is a classic two-stage algorithm known for high detection accuracy but faces challenges such as slow speed, training complexity, and optimization \cite{deng2018}. One-stage detectors, like the YOLO series \cite{redmon2016yolo, redmon2017yolo9000} and SSD \cite{liu2016ssd}, use a single neural network to predict box coordinates and categories in one pass. Consequently, single-stage networks excel in applications where speed is crucial. However, they sacrifice some accuracy. Despite advancements in speed, these methods struggle with accuracy due to the multi-scale nature of objects and the prevalence of small objects in UAV and traffic scenes. 

Recent research has focused on improving small object detection in UAV aerial photography and traffic scenarios, which is challenging due to their lower resolution and less distinct visual characteristics compared to larger objects. Studies have explored diverse backbone architectures to enhance feature representation, reduce false positives, and extract relevant features from complex backgrounds in UAV imagery.

Liu et al. \cite{liu2020uavyolo} introduced a model for small target detection in UAV images, addressing leakage and false positives by integrating ResNet units and optimizing convolutional operations to expand the network's receptive field. Liu et al. \cite{liu2022uav} proposed CBSSD, a specialized detector for small object detection in UAV traffic images. By integrating ResNet50's lower-level features with VGG16, CBSSD improves feature representation, enhances object recognition accuracy, and reduces false positives under challenging lighting conditions. Additionally, Liu et al. \cite{liu2020small} utilized Multi-branch Parallel Feature Pyramid Networks (MPFPN) and SSAM for detecting small objects in UAV images. Their approach enhances feature extraction through MPFPN for deep layer detail recovery, while SSAM reduces background noise, significantly boosting accuracy. Experimental results on the VisDrone-DET dataset \cite{Zhu2021} showcase their method's competitive performance. 

Adaptations and optimizations within the YOLO framework have also been explored to address challenges in small object detection. Lai et al. \cite{lai2023stcyolo} introduced STC-YOLO, a specialized variant of YOLOv5 designed for challenging traffic sign detection. Their improvements include refined down-sampling, a dedicated small object detection layer, and a CNN-based feature extraction module with multi-head attention. STC-YOLO demonstrated a significant 9.3\% improvement in mean Average Precision (mAP) compared to YOLOv5 on benchmark datasets.

Further enhancements have been made to YOLOv8, focusing on improving backbone architectures, integrating attention mechanisms to focus on relevant features and suppress irrelevant ones, and modifying loss functions. Shen et al. \cite{shen2023dsyolov8} introduced DS-YOLOv8 to enhance small object detection within images by integrating Deformable Convolution C2f (DCN\_C2f) and Self-Calibrating Shuffle Attention (SC\_SA) for adaptive feature adjustment, alongside Wise-IoU \cite{tong2023wiseiou} and position regression loss to boost performance. Experimental results across diverse datasets show significant enhancements in $mAP_{0.5}$. Wang et al. \cite{wang2023uav} improved YOLOv8 for UAV aerial photography with a BiFormer attention mechanism for focusing on important information and FFNB for effective multiscale feature fusion. This resulted in a significant 7.7\% increase in mean detection accuracy over baseline models, surpassing widely-used alternatives in detecting small objects. This marks substantial progress in UAV object detection, though it necessitates further optimization due to increased computational complexity from additional detection layers. 

Wang et al. \cite{wang2024remotesensing} improved YOLOv8 for detecting targets in remote sensing images, focusing on complex backgrounds and diverse small targets. They introduced a small target detection layer and incorporated a C2f-E structure using the EMA attention module. Experimental results on the DOTAv1.0 dataset \cite{xia2018dota} demonstrate a notable 1.3\% increase in $mAP_{0.5}$ to 82.7\%, highlighting significant advancements in target detection accuracy. However, their approach introduces increased computational complexity. Xu et al. \cite{xu2024yolov8mpeb} introduced YOLOv8-MPEB, specialized for small target detection in UAV images, addressing scale variations and complex scenes. Enhancements include replacing CSPDarknet53 \cite{redmon2018yolov3} with MobileNetV3 for efficiency, integrating Efficient Multi-Scale Attention in C2f for better feature extraction, and incorporating Bidirectional Feature Pyramid Network (BiFPN) \cite{tan2020efficientdet} in the Neck segment for enhanced adaptability. Experimental results on a custom dataset demonstrate YOLOv8-MPEB achieving a 91.9\% mAP, a 2.2\% improvement over standard YOLOv8, while reducing parameters by 34\% and model size by 32\%. However, accurately detecting dense small targets remains a challenge.

Despite advancements in the reviewed studies, small object detection methods still face challenges in UAV aerial photography and traffic scenarios. These methods primarily focus on feature fusion but often neglect inner block connections. In contrast, our approach integrates an optimized GFPN, inspired by Efficient-RepGFPN, into YOLOv8. This enhancement incorporates skip connections and queen fusion structures to improve efficacy without significantly increasing computational complexity or latency. Additionally, the introduced C2f-EMA module enhances feature extraction by redistributing feature weights using the EMA attention mechanism. Unlike other attention mechanisms, it overcomes limitations such as neglecting interactions among spatial details and the limited receptive field of 1x1 kernel convolution, which limits local cross-channel interaction and contextual information modeling.

Furthermore, our method avoids the enlargement issues common in other bounding box regression methods. The used PIoU loss function effectively guides anchor boxes during training, resulting in faster convergence and demonstrating its effectiveness. While existing methods perform well in controlled datasets, they often struggle to generalize across diverse environments and dynamic lighting conditions in real-world settings. In this paper, we experiment with real-world traffic scenes captured by building-mounted cameras, assessing diverse environments, lighting conditions, and dynamic scenarios such as nighttime and crowded scenes. This challenges the generalization capabilities of small object detection method beyond controlled datasets.

\section{Introduction of YOLOv8 Detection Network}

\begin{figure}
    \centering
    \includegraphics[width=1\linewidth]{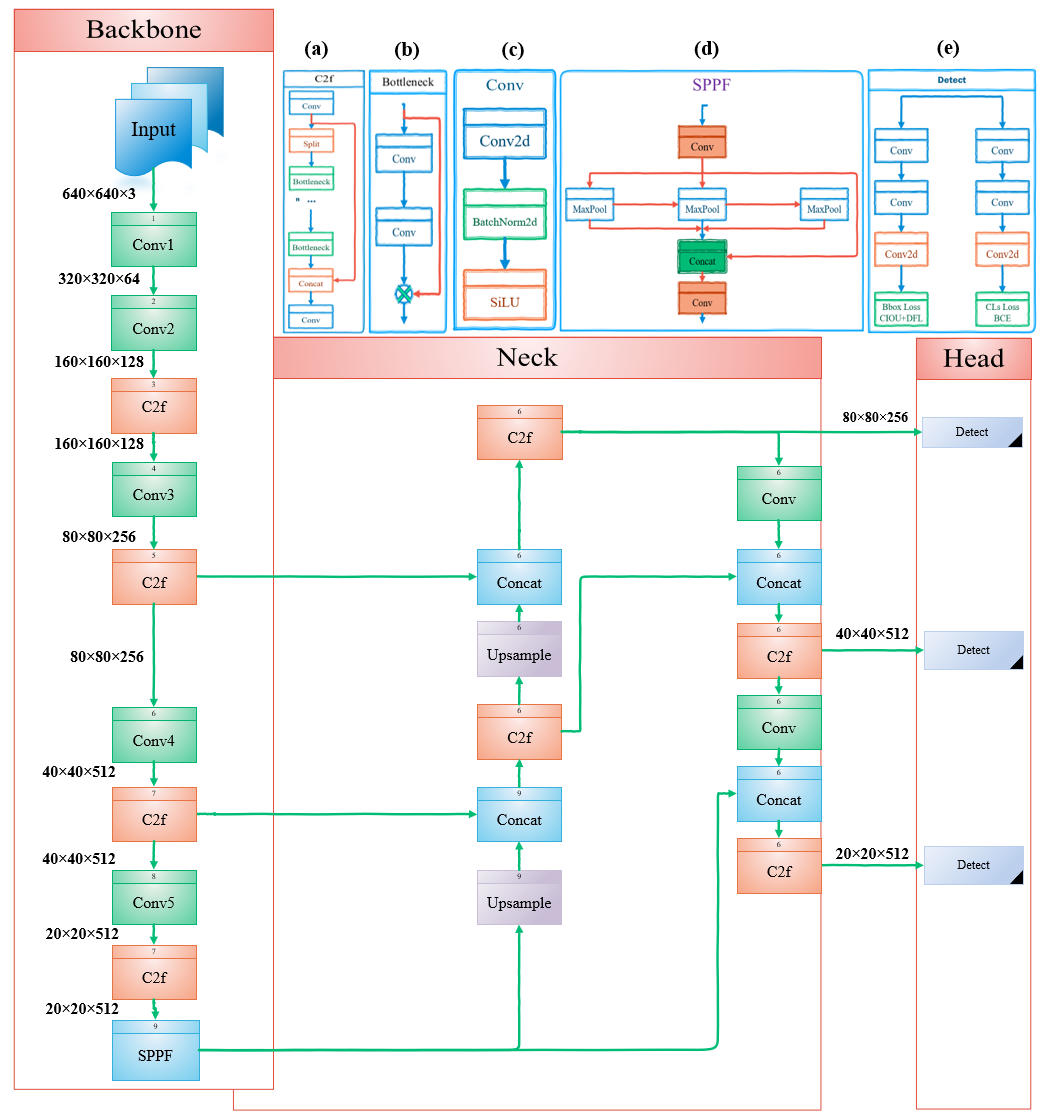}
    \caption{The network structure of YOLOv8.}
    \label{fig:YOLOv8}
\end{figure}

As shown in Figure \ref{fig:YOLOv8}, the YOLOv8 architecture consists of three main elements: the backbone, neck, and detection layers. Each of these components will be introduced in the subsequent sections.

\subsection{Backbone Layer}

The architecture of YOLOv8 is based on the CSPDarknet53 \cite{redmon2018yolov3} backbone, employing five downsampling stages to extract distinct scale features. It improves information flow and stays lightweight by using the C2f module instead of the Cross Stage Partial (CSP) module \cite{wang2020cspnet}. The C2f module includes dense and residual structures for better gradient flow and feature representation. The backbone also includes the Spatial Pyramid Pooling Fast (SPPF) module, which captures features at multiple scales to boost detection performance. The SPPF layer reduces computational complexity and latency while optimizing feature extraction \cite{he2015spatial}.

\subsection{Neck Layer}

For multi-scale feature fusion, YOLOv8's neck uses Feature Pyramid Network (FPN) \cite{lin2017fpn} and Path Aggregation Networks (PANet) \cite{liu2018pan}. FPN enhances hierarchical feature fusion, improving object detection across various scales through a top-down pathway, while PANet enhances feature representation and information reuse with a bottom-up pathway, though it increases computational cost. Combining FPN-PANet structures and C2f modules integrates feature maps of various scales, merging both shallow and deep information.

\subsection{ Detection Head Layer}

YOLOv8, a state-of-the-art object detection model, enhances accuracy and robustness by using the Task-Aligned Assigner \cite{feng2021tood} instead of traditional anchors. This assigner dynamically categorizes samples as positives or negatives, refining the model's ability to detect objects accurately. The detection head features a decoupled structure with separate branches for object classification and bounding box regression. For classification, it employs binary cross-entropy loss (BCE Loss). For regression, it uses a combination of distribution focal loss (DFL) \cite{li2020generalizedfocalloss} and Complete Intersection over Union (CIoU) \cite{zheng2020distanceiou} loss. These efficient loss functions are crucial for precise object localization, further boosting the model's performance.

Bounding box loss functions aim to accurately localize objects by penalizing differences between predicted and ground truth bounding boxes. IoU-based loss functions \cite{yu2016unitbox} are crucial for bounding box regression in the detection layer. IoU Loss measures the overlap between predicted and ground truth boxes by comparing the ratio of their intersection area to their union area. However, its gradient diminishes when there is no overlap, making it less effective in such cases. Various IoU-based loss functions have been developed with different methodologies and constraints. CIoU, used in YOLOv8, minimizes the normalized distance between the center points of predicted and ground truth boxes and includes an aspect ratio penalty term. This approach improves convergence speed and overall performance.

\section{Method}

\begin{figure}
    \centering
    \includegraphics[width=1\linewidth]{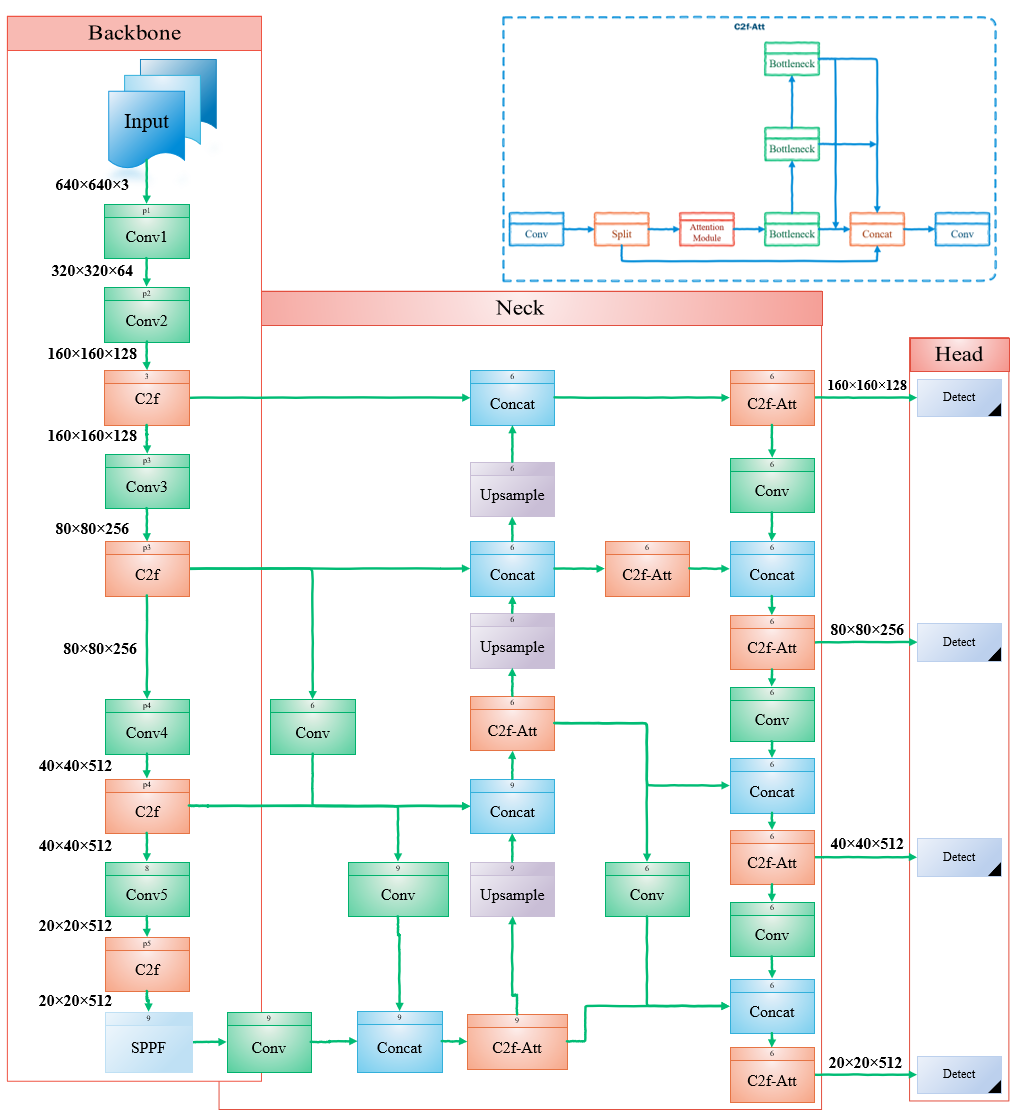}
    \caption{Proposed improved YOLOv8 for small object detection}
    \label{fig:enhanced-model}
\end{figure}
In this section, we introduce three pivotal enhancements to improve small object detection in our study. First, we enhance feature fusion within the YOLOv8 architecture's neck to better retain crucial spatial details typically filtered out by shallower layers. This modification aims to mitigate information loss, especially for smaller objects overshadowed by larger ones during feature extraction. Second, we propose the C2f-EMA module, integrating an EMA attention mechanism to prioritize relevant features and spatial details across different channels. This method enhances feature extraction efficiency by redistributing feature weights effectively. Finally, we use PIoU as an improved bounding box regression metric, replacing CIoU. PIoU incorporates a penalty term that minimizes the Euclidean distance between corresponding corners of predicted and ground truth boxes, offering a more intuitive measure of similarity and stability in box regression tasks. These methods contribute to enhancing the accuracy and robustness of our small object detection framework. The enhanced structure depicted in Figure \ref{fig:enhanced-model} is utilized in this paper.

\subsection{Improved GFPN for Multilevel Feature Integration}

In YOLOv8, crucial spatial details are primarily encoded in the network's shallower layers. However, these layers often filter out less prominent details, leading to significant data loss for small object detection. Additionally, smaller objects may be overshadowed by larger ones during feature extraction, resulting in a gradual loss of information and the potential disappearance of relevant details. To address these challenges, this study introduces an enhanced feature fusion method in the neck of the YOLOv8 architecture. This method focuses on preserving and effectively utilizing important information from the shallower layers, thereby enhancing overall detection accuracy, especially for small objects.

The FPN merges features of different resolutions extracted from a backbone network. It begins with the highest-resolution feature map and progressively combines features from higher to lower resolutions using a top-down approach. PAFPN improves FPN by adding a bottom-up approach that enhances bidirectional information flow. It merges features from lower to higher network layers, prioritizing spatial detail preservation, even with increased computational demands.

The BiFPN \cite{tan2020efficientdet} enhances object detection by integrating features across different resolutions bidirectionally, using both bottom-up and top-down pathways. This method optimizes multi-scale feature utilization, simplifies the network by reducing computational complexity, and includes skip connections at each level. These connections allow for adaptable use of input features, enhancing feature fusion across scales and details \cite{tan2020efficientdet}. However, deep stacking of BiFPN blocks may cause gradient vanishing during training, potentially affecting overall network performance \cite{kang2023bgfyolo}.

Prior methods focused mainly on combining features without considering inner block connections. In contrast, GFPN introduces skip connections and queen fusion structures. It employs skip-layer and cross-scale connections to enhance feature combination. GFPN implements skip connections in two forms: $\log_2(n)$-link and dense-link.

The $\log_2(n)$-link method optimizes information transmission by allowing the $l^{\text{th}}$ layer at level $k$ to receive feature maps from up to $\log_2(l) + 1$ preceding layers. These skip connections help mitigate gradient vanishing during back-propagation by extending the shortest gradient distance from one layer to approximately $1 + \log_2(n)$ layers \cite{jiang2022giraffedet}. This extension facilitates more effective gradient propagation over longer distances, potentially enhancing the scalability of deeper networks.

\begin{figure}
    \centering
    \includegraphics[width=1\linewidth]{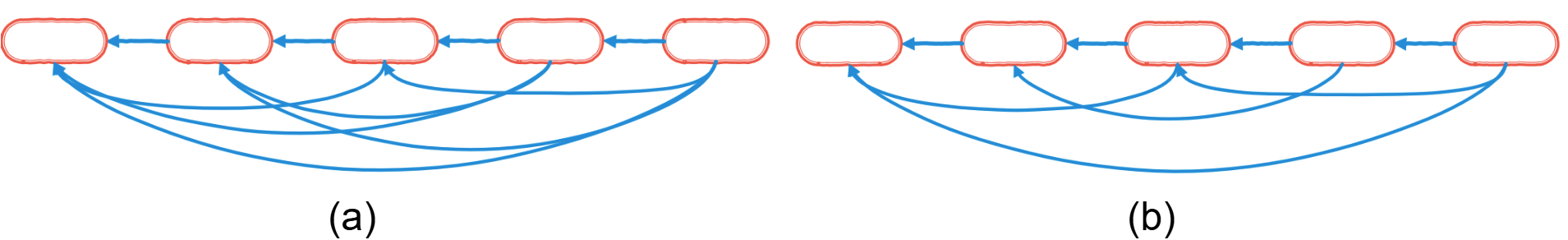}
    \caption{skip-layer links: (a) dense-link: concatenates features from all preceding layers; (b) $\log_2n$-link: concatenates features from up to $\log_2(l)+1$ layers at each level.}
    \label{fig:gpfnLinks}
\end{figure}

In contrast, the dense-link method ensures that each scale feature $P_{k}^{l}$ at level $k$ receives feature maps from all preceding layers up to the $l^{\text{th}}$ layer. This promotes robust information flow and integration of features across multiple scales. In GFPN, this structure facilitates seamless connectivity between layers, enhancing feature reuse and improving the network's efficiency in tasks like object detection. During back-propagation, the dense connectivity in GFPN supports efficient transmission of feature information across the network hierarchy. Dense-link and $\log_2(n)$-link configurations are illustrated in Figure \ref{fig:gpfnLinks}.

\begin{figure}
    \centering
    \includegraphics[width=1\linewidth]{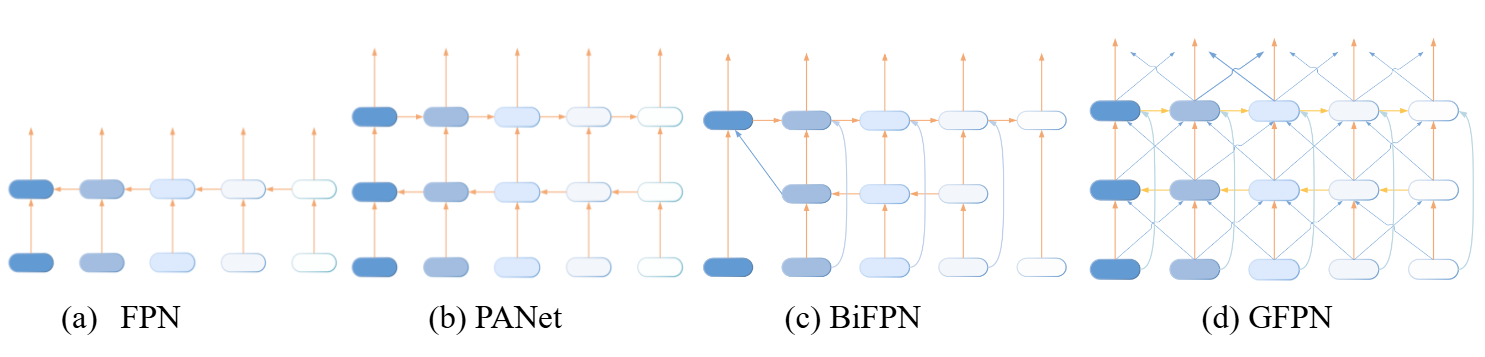}
    \caption{Different Feature Pyramid Network designs: (a) FPN uses a top-down strategy; (b) PANet enhances FPN with a bottom-up pathway; (c) BiFPN integrates cross-scale pathways bidirectionally; (d) GFPN includes a queen-fusion style pathway and skip-layer connections.}
    \label{fig:FPNevoulution}
\end{figure}

Another significant improvement in GFPN is the Queen-Fusion module, which facilitates cross-scale connections to enhance adaptability to multi-scale variations. This module utilizes a 3x3 convolution to merge features across different scales, gathering input features from diagonally adjacent nodes above and below to minimize information loss during feature fusion. Implementing this approach enhances the network's capability to handle multi-scale variations, potentially improving overall performance robustness. Figure \ref{fig:FPNevoulution} illustrates various methods for integrating features across different layers, including FPN, PANet, BiFPN, and GFPN.

In YOLOv8, integrating PAFPN with C2f modules effectively combines feature maps across scales, thereby enhancing object detection capabilities. This study aims to enhance YOLOv8's small object detection using advanced feature fusion techniques. However, replacing PAFPN with GFPN in YOLOv8 improves precision while introducing higher latency compared to the PAFPN-based model.

\begin{figure}
    \centering
    \includegraphics[width=1\linewidth]{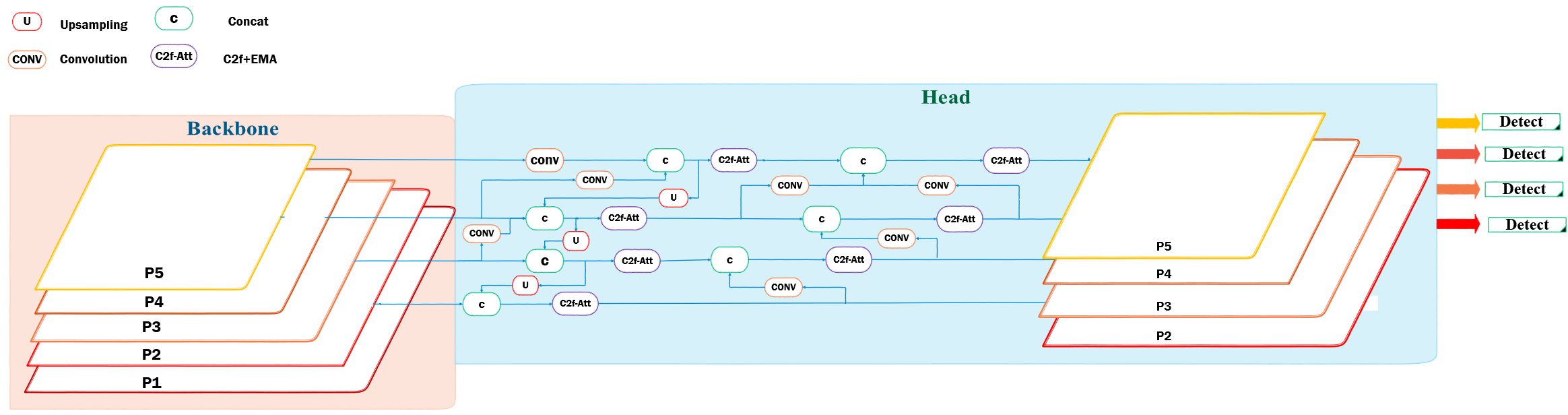}
    \caption{Enhanced and efficient GPFN structure}
    \label{fig:proposed-net}
\end{figure}

This paper introduces an enhanced and efficient GFPN, depicted in Figure \ref{fig:proposed-net}, inspired by Efficient-RepGFPN \cite{xu2022damoyolo}. By integrating it into YOLOv8, the model achieves superior performance without significantly increasing computational complexity or latency. Efficient-RepGFPN simplifies complexity by parameterizing and eliminating additional upsampling operations in queen-fusion. Furthermore, it upgrades the feature fusion module to CSPNet, enhancing the merging of features across different scales \cite{xu2022damoyolo}.

In the feature fusion block of the GFPN architecture, we replace the conventional 3x3 convolution-based feature fusion with C2f-EMA, incorporating an attention mechanism. This module merges high-level semantic features with low-level spatial details, thereby enhancing the representation and detection accuracy of small objects. These modifications maintain GFPN's ability to improve feature interaction and efficiency by effectively managing both types of information in the neck section. Inspired by Efficient-RepGFPN, we also reparametrize and eliminate additional upsampling operations in queen-fusion. Ultimately, these enhancements improve the efficiency and effectiveness of YOLOv8 for object detection tasks without significantly increasing computational complexity or latency.

We enhance the network's capability by adding a detection layer, which involves integrating feature maps from P2 alongside those from P3 to P5, which are used in YOLOv8. This enhancement significantly improves the network's ability to detect small objects. As depicted in Figure \ref{fig:proposed-net}, P2 with a resolution of 320x320 plays a crucial role in preserving finer spatial details essential for improving small object detection. Additionally, a detection head is introduced, prompting the network structure to focus more on features related to small objects.

This approach not only provides higher-resolution details but also enhances feature fusion, offers comprehensive contextual information, and enables precise localization. By leveraging features from both fine and coarse scales simultaneously, the network achieves accurate detection of small objects, effectively capturing finer details. The enhanced architectural design illustrated in Figure \ref{fig:proposed-net} is implemented and evaluated in this study.

\subsection{Embedding Efficient Multi-scale Attention Mechanism in C2f}

The C2f module in YOLOv8 enhances gradient flow and detection accuracy by dynamically adjusting channel numbers through split and concatenate operations, optimizing feature extraction while managing computational complexity \cite{wang2024remote}. It incorporates convolutional and residual structures to deepen network training and address the vanishing gradient problem, thereby improving feature extraction.

\begin{figure}
    \centering
    \includegraphics[width=1\linewidth]{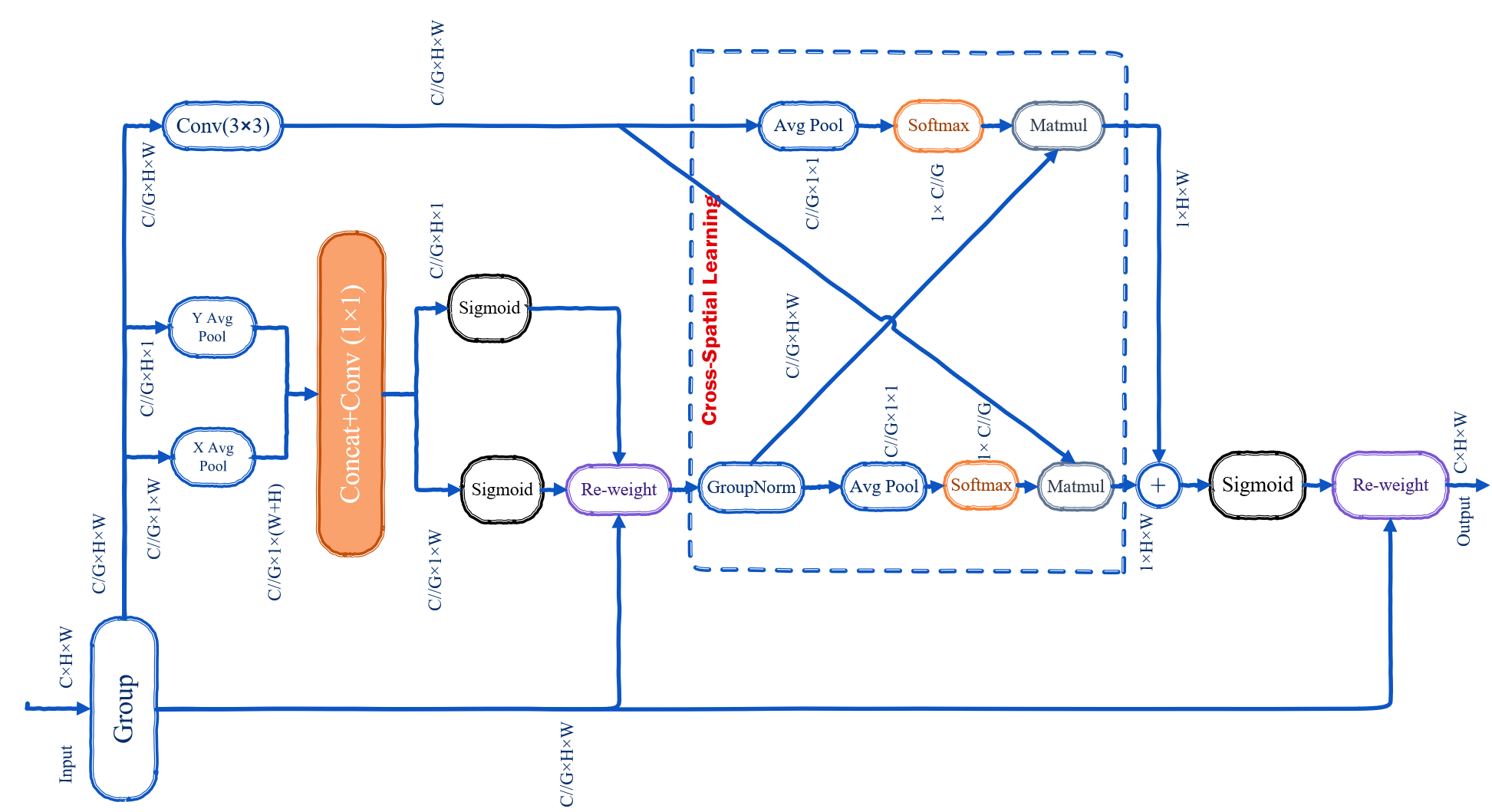}
    \caption{Efficient Multi-scale Attention Mechanism}
    \label{fig:EMA}
\end{figure}

The C2f-EMA module introduced in this paper enhances feature extraction by redistributing feature weights using the EMA attention mechanism. This mechanism prioritizes relevant features and spatial details across different channels within the image. Figure \ref{fig:EMA} illustrates the EMA structure, which partitions input features into groups, processes them through parallel subnetworks, and integrates them with advanced aggregation techniques. This enhancement significantly boosts the representation of small, challenging objects and enhances the efficiency of the network backbone.

The EMA module employs feature grouping to partition the input feature map \( X \) along the channel dimension into \( G \) sub-features, denoted as \( X = [X_1, X_2, \ldots, X_G] \), where each \( X_i \in \mathbb{R}^{\frac{C}{G} \times H \times W} \). This approach enables specialized feature extraction and representation by allowing the network to learn different semantics or characteristics within each group. Additionally, it optimizes CNNs by reducing computation.

The EMA module uses a Parallel Subnetworks approach to efficiently capture multi-scale spatial information and cross-channel dependencies. It features two parallel branches: the 1x1 Branch, with two routes, and the 3x3 Branch, with one route. In the 1x1 Branch, each route employs 1D global average pooling to encode channel information along the horizontal and vertical spatial directions. These operations produce two encoded feature vectors representing global information, which are then concatenated along the height direction. A 1x1 convolution layer is subsequently applied to the concatenated output to maintain channel integrity and capture cross-channel interactions by blending information across different channels. The outputs are split into two vectors, refined using non-linear Sigmoid functions to adjust attention weights in a 2D Binomial distribution. Channel-wise attention maps are then combined through multiplication within each group, enhancing interactive features across channels.

EMA differs from traditional attention methods by addressing issues such as neglecting interactions among spatial details and the limited scope of 1x1 kernel convolution, accomplished through the inclusion of a 3x3 convolution branch. This branch uses a single route with a 3x3 convolution kernel to capture multi-scale spatial information. Additionally, the output of the 1x1 branch undergoes 2D global average pooling to encode global spatial information. The pooled output integrates with the transformed output from the 3x3 branch, aligning dimensions to enhance feature aggregation through both spatial information sources.

Unlike traditional attention methods that use basic averaging, EMA integrates attention maps from parallel subnetworks using a cross-spatial learning approach. It uses matrix dot product operations to capture relationships between individual pixels, enriching global context across all pixels. Specifically, the EMA module combines global and local spatial information from its parallel 1x1 and 3x3 branches to enhance feature representation. This approach effectively captures long-range dependencies and multi-scale spatial details, improving overall feature aggregation.

SoftMax is then applied to the outputs to generate 2D Gaussian maps, highlighting relevant features and modeling long-range dependencies. This process is repeated for the second spatial attention map, using 2D global average pooling and Sigmoid functions to preserve precise spatial positional information. Finally, feature maps obtained from spatial attention weights within each group are aggregated. The resulting output feature map retains the original input size, ensuring efficiency and effectiveness for integration into architectures. The final output is a redistributed feature map that captures pixel-level pairwise relationships and highlights the global context across all pixels and channels, allocating higher weights to more relevant features and spatial details.

\begin{figure}
    \centering
    \includegraphics[width=1\linewidth]{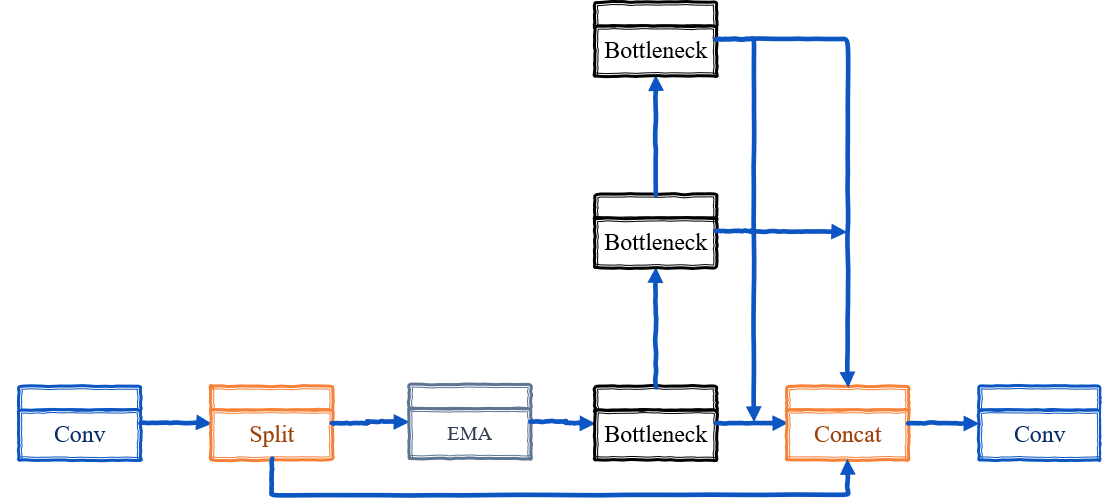}
    \caption{C2f-EMA}
    \label{fig:c2f-EMA}
\end{figure}

In this paper, we introduce the C2f-EMA as a replacement for C2f, redistributing the feature map using the EMA structure to assign higher weights to more relevant features and spatial details within images. This enhancement aims to improve detection performance, especially for small objects with very fine details due to their size. C2f-EMA includes initial convolution, split function, EMA module, and parallel processing, collectively enhancing the network's overall performance. As shown in Figure \ref{fig:c2f-EMA}, this mechanism operates within the second residual block of the C2f.

\subsection{Improved Bounding Box Loss Function}

Bounding box loss functions penalize discrepancies between predicted and ground truth bounding box parameters to enhance object localization. IoU-based loss functions are essential for this purpose, measuring overlap as the ratio of intersection over union. However, their effectiveness diminishes when there is no overlap, resulting in negligible gradients. Several IoU-based loss functions have been developed to address this limitation, each presenting unique approaches and specific limitations. CIoU, implemented in YOLOv8, considers the distance between box centers and differences in aspect ratios. This refinement enhances convergence speed and overall performance. However, the aspect ratio penalty term in CIoU may not sufficiently account for size variations between boxes of the same aspect ratio but different dimensions. Moreover, CIoU involves computationally intensive inverse trigonometric functions, which could pose drawbacks in real-time applications. The equation for CIoU is presented in Equation \ref{eqCIOU}:

\begin{equation} \label{eqCIOU}
\begin{aligned}
L_{\text{CIOU}} &= L_{\text{IOU}} + \frac{d^2}{c^2} + v, \\
v &= \frac{4}{\pi^2} \left( \arctan \left( \frac{w_{\text{gt}}}{h_{\text{gt}}} \right) - \arctan \left( \frac{w}{h} \right) \right)^2
\end{aligned}
\end{equation}

Where d represents the Euclidean distance between the center points of the predicted and ground truth bounding boxes. Additionally, c represents a normalization factor, typically representing the diagonal length of the smallest enclosing box that contains both the predicted and ground truth bounding boxes.  \( v \) represents the aspect ratio penalty, which accounts for discrepancies in aspect ratios between the predicted and ground truth boxes. \((w, h)\) and \((w_{\text{gt}}, h_{\text{gt}})\) represent the width and height of the predicted and  ground truth bounding box, respectively.

Efficient Intersection over Union (EIoU) adjusts CIoU by using distinct penalty terms for width and height rather than a shared aspect ratio penalty, aiming for more precise measurement of differences between anchor box and target box dimensions. Equation \ref{eqEIOU} presents the EIoU formula.

\begin{equation} \label{eqEIOU}
L_{\text{EIOU}} = L_{\text{IOU}} + \frac{d^2}{c^2} + \frac{(w_{\text{pred}} - w_{\text{gt}})^2}{w_c^2} + \frac{(h_{\text{pred}} - h_{\text{gt}})^2}{h_c^2}
\end{equation}

where \( w_c \) and \( h_c \) represent the width and height of the smallest enclosing bounding box, respectively. Despite addressing size discrepancies, EIoU encounters challenges such as anchor box enlargement during regression and slow convergence. This issue is critical in object detection models that use IoU-based loss functions, where optimization may inadvertently enlarge anchor boxes instead of precisely converging them to target sizes, thereby reducing localization precision \cite{liu2024powerfuliou}. CIoU and EIoU losses use the term \( R_D = \frac{d^2}{c^2} \), where \( d \) is the diagonal length of the intersection between the anchor and target boxes, and \( c \) is the diagonal length of the smallest enclosing box covering both. The gradient of \( R_D \) with respect to \( d \) is \(\frac{2d}{c^2}\), meaning \( R_D \) decreases as \( c \) increases. The problem is when the boxes do not overlap, enlarging the anchor box increases \( c \), reducing \( R_D \) and thus lowering the CIoU and EIoU losses without improving the overlap. Using \( c \) as the denominator in the penalty term is flawed, allowing loss reduction through anchor box size manipulation rather than overlap improvement, indicating the need for a revised penalty term to better handle non-overlapping boxes \cite{liu2024powerfuliou}.

Wise Intersection over Union (WIoU) \cite{tong2023wiseiou} introduces a dynamic, non-monotonic focusing mechanism in bounding box regression. This mechanism prioritizes anchor boxes with moderate quality and reduces harmful gradients from low-quality examples. WIoU uses aspect ratio and the distance between predicted and ground truth boxes as penalty terms. It evaluates anchor box quality dynamically by comparing each box's quality to the average quality of all boxes in the batch, giving more attention to those with moderate quality. The WIoU calculation is given by Equation \ref{eqWIOU}.

\begin{equation}\label{eqWIOU}
\begin{aligned}
L_{\text{WIoUv3}} &= \frac{\beta}{\delta \alpha^{\beta - \delta}} \cdot 
e^{\left(\frac{(x - x_{gt})^2 + (y - y_{gt})^2}{w_{gt}^2 + h_{gt}^2}\right)}, \\
\beta &= \frac{{L}_{\text{IOU}}^*}{\overline{L}_{\text{IOU}}}
\end{aligned}
\end{equation}

The non-monotonic attention function of WIoU is denoted by \( \beta \), while \( \delta \) and \( \alpha \) serve as hyper-parameters that regulate its gradient. The operation \( * \) denotes the detach operation, and \( L_{\text{IOU}} \) indicates the average \( L_{\text{IOU}} \) value across all anchor boxes within a batch. WIoU introduces attention-based predicted box loss and focusing coefficients. However, it relies on multiple hyper-parameters, posing challenges in optimization for diverse datasets.

\begin{figure}
    \centering
    \includegraphics[width=1\linewidth]{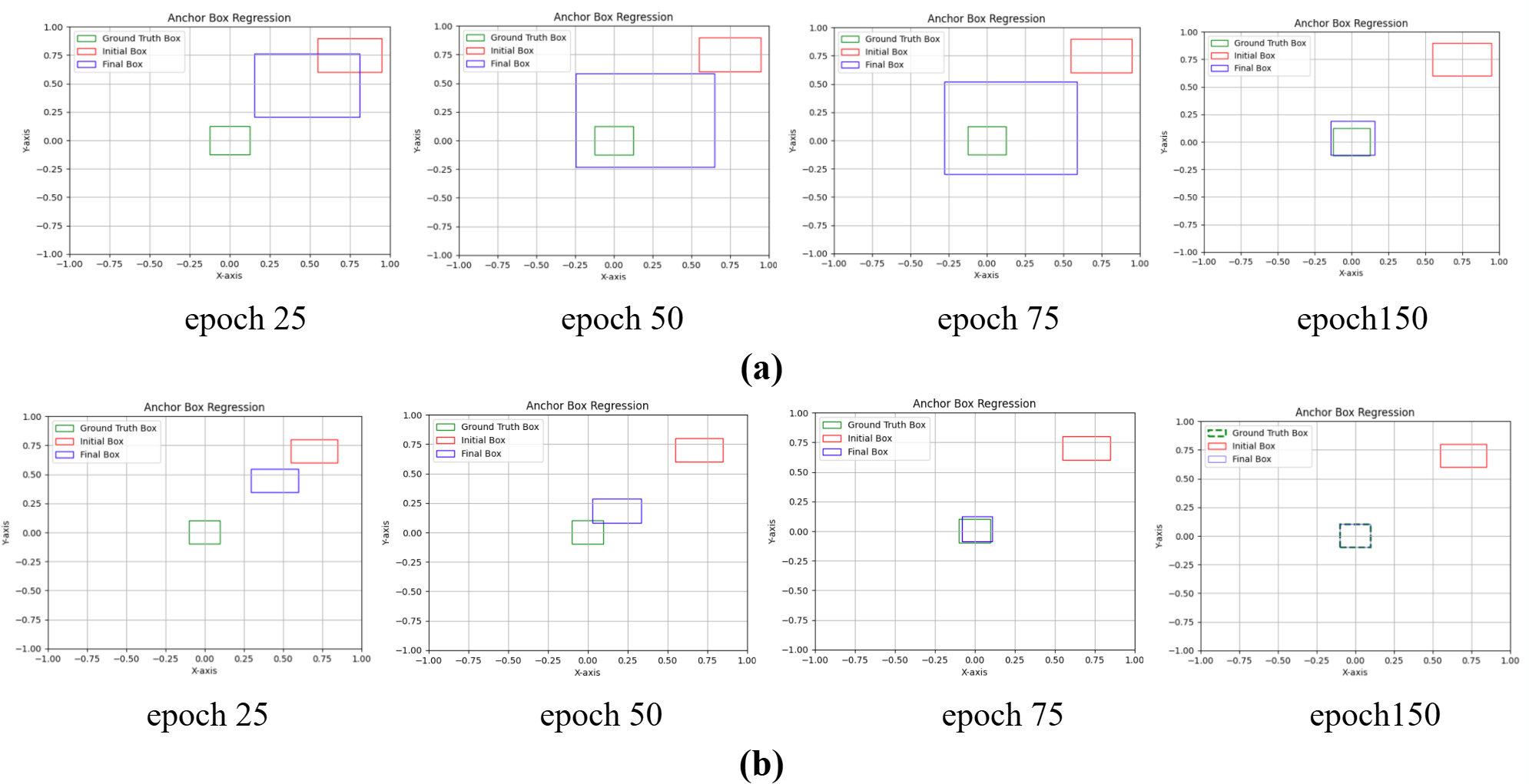}
    \caption{Anchor box regression process guided by (a) Complete IoU-based loss function (CIoU) (b)  penalty term in Powerful-IoU (PIoU) loss function without attention function.}
    \label{fig:enlargment}
\end{figure}

We use PIoU \cite{liu2024powerfuliou} as a replacement for CIoU in the original network. The details of the PIoU method are outlined in Algorithm 1. The penalty term in PIoU enhanced bounding box regression by minimizing the Euclidean distance between corresponding corners of predicted and ground truth boxes. This approach offers a more intuitive measure of similarity and proves effective for both overlapping and non-overlapping boxes. Unlike traditional IoU-based loss functions, PIoU mitigates the issue of box enlargement, ensuring precise and stable box regression. Simulated results in Figure \ref{fig:enlargment} demonstrate its effectiveness. Figure \ref{fig:enlargment} illustrates an experiment evaluating anchor box regression using various loss functions. The CIoU loss function exhibits continuous enlargement of anchor boxes from epochs 25 to 75 and fails to achieve full convergence to the ground truth anchor box by epoch 150. In contrast, the anchor box guided solely by the penalty term in the PIoU loss function, without consideration of the attention function, does not show enlargement issues during training, as observed in epochs 25 and 75. By epoch 75, it demonstrates almost complete convergence to the ground truth bounding box, reaching a perfect fit by epoch 150.
PIoU loss uses a non-monotonic attention layer to enhance focus on medium and high-quality anchor boxes. By prioritizing moderate-quality stages in anchor box regression, PIoU improves object detector performance. The non-monotonic attention function \( u(\lambda q) \), controlled by the parameter \( \lambda \). PIoU simplifies the tuning process by requiring only one hyperparameter. Replace the penalty factor \(P\) with \(q\), which indicates anchor box quality on a scale from 0 to 1. When \(q = 1\) (meaning \(P = 0\)), the anchor box perfectly aligns with the target box. As \(P\) increases, \(q\) decreases, signifying lower-quality anchor boxes.

\begin{algorithm}
\caption{: PIoU Bounding Box Regression}
\SetAlgoLined
\textbf{Input:}\\
- Two arbitrary convex shapes: \(A, B \subseteq \mathbb{R}^n\) \\
- Width and height of input image: \(w, h\) \\
- Width and height of ground truth box: \(w_{\text{gt}},h_{\text{gt}}\)\\
- Coordinates of bounding box 1: \(b1_{x1}, b1_{x2}, b1_{y1}, b1_{y2}\) \\
- Coordinates of bounding box 2: \(b2_{x1}, b2_{x2}, b2_{y1}, b2_{y2}\) \\
- \(\text{IoU}\) (Intersection over Union) \\
\textbf{Output:}\\
- \(\text{ Powerful-IoU}\) \\
\textbf{Steps:}
\begin{enumerate}
    \item Calculate Absolute Differences for Widths:
    \begin{itemize}
        \item \(dw1 = | \min(b1_{x2} - b1_{x1}) - \min(b2_{x2} - b2_{x1}) |\)
        \item \(dw2 = | \min(b1_{x2} - b1_{x1}) - \min(b2_{x2} - b2_{x1}) |\)
    \end{itemize}
        \item Calculate Absolute Differences for Heights:
    \begin{itemize}
        \item \(dh1 = | \min(b1_{y2} - b1_{y1}) - \min(b2_{y2} - b2_{y1}) |\)
        \item \(dh2 = | \min(b1_{y2} - b1_{y1}) - \min(b2_{y2} - b2_{y1}) |\)
    \end{itemize}
        \item Compute Parameter \(P\):
    \[ P = \frac{1}{4} (\frac{dw1 + dw2}{w_{\text{gt}}} + \frac{dh1 + dh2}{ h_{\text{gt}}}) \]
        \item Calculate Custom Loss \(L\):
    \[ L = 1 - \text{IoU} - e^{-P^2 } \]
        \item Calculate the Focal Loss by Adding an Attention Layer:

    \[ q = e^{-P}, \quad q \in (0, 1] \]
    \[ u(x) = 3x \cdot e^{-x^2} \]
     \[ L_{PIoU} = u(\lambda q) \cdot L\]
    \[ L_{PIoU} = 3 \cdot (\lambda q) \cdot e^{-(\lambda q)^2} \cdot L \]
\end{enumerate}
\end{algorithm}

\section{Results}

This section begins with an introduction to the dataset utilized in this paper, followed by detailing the experimental environment and training strategy. It further outlines the evaluation metrics employed to assess the model's performance. The effectiveness of the proposed approach is then demonstrated through a comparative analysis with state-of-the-art models, using YOLOv8 as the baseline. Furthermore, the section includes an evaluation of the model's performance in challenging real-world scenarios, such as detecting distant objects and small objects positioned far from the camera.

\subsection{Dataset}

\begin{figure}
    \centering
    \includegraphics[width=1\linewidth]{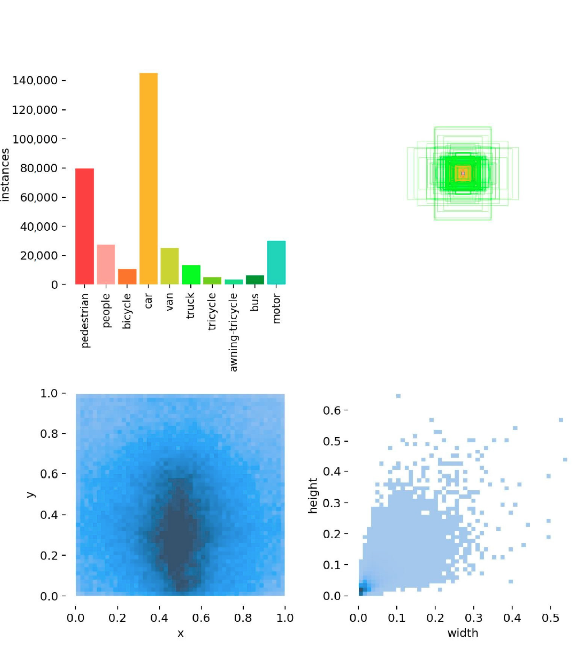}
    \caption{Information regarding the manual annotation process for objects in the VisDrone2019 dataset}
    \label{fig:VisDrone}
\end{figure}

The VisDrone2019 dataset \cite{Zhu2021}, a prominent collection of UAV aerial photography, was developed by Tianjin University's Lab of Machine Learning and Data Mining in collaboration with the AISKYEYE data mining team. It comprises 288 video clips totaling 261,908 frames and 10,209 static images. These visuals were captured using various drone-mounted cameras, showcasing diverse scenarios across more than a dozen cities throughout China. The dataset is exceptionally rich, featuring a wide range of geographic locations, environmental settings, and object types. Geographically, the dataset covers footage from 14 different cities across China, offering a comprehensive spectrum of scenes from urban to rural landscapes. It includes a diverse array of objects such as pedestrians, cars, bicycles, and more. Additionally, the dataset spans various population densities, ranging from sparse to densely crowded areas, and captures images under different lighting conditions, including both daytime and nighttime scenes. One distinguishing feature of the VisDrone2019 dataset is its inclusion of numerous small objects of varying sizes, depicted from different angles and within various scenes. This diversity increases the dataset's complexity and difficulty compared to other computer vision datasets. Figure \ref{fig:VisDrone} illustrates the process of manually annotating objects in the VisDrone2019 dataset.

\subsection{Experimental Environment and Training Strategies}

In this study, YOLOv8s was selected as the baseline model for investigation and further enhancements. The model was trained on the VisDrone dataset using an NVIDIA RTX A6000 GPU (48 GB) on Linux, utilizing PyTorch 2.2.1 and CUDA 12.1. Training involved optimizing key parameters, running for 200 epochs with the Stochastic Gradient Descent (SGD) optimizer \cite{loshchilov2017sgdr} set to a momentum of 0.932. The initial learning rate started at 0.01 and decayed gradually to 0.0001. A batch size of 32 was chosen for efficient memory usage and stable training, with input images resized to 640x640 pixels. A weight decay of 0.0005 was also applied to prevent overfitting and improve model generalization.

\subsection{Evaluation metrics}

To assess the detection performance of our enhanced model, we utilize several evaluation metrics: precision, recall, mAP${0.5}$, mAP${0.5:0.95}$, and the number of model parameters. The specific formulas for these metrics are provided in this section.

\textbf{Precision:} is the metric that represents the ratio of true positives to the total predicted positives, as defined by Equation \ref{Precision}:
\begin{equation} \label{Precision}
\text{Precision} = \frac{\text{TP}}{\text{TP} + \text{FP}}
\end{equation}
True Positives (TP) is the number of instances where the model accurately predicts a positive instance. False Positives (FP) is the number of instances where the model incorrectly predicts a positive instance. False Negatives (FN) is the number of instances where the model fails to predict a positive instance.

\textbf{Recall:} Measures the ratio of correctly predicted positive samples to all actual positive samples, as defined by Equation \ref{Recall}:
\begin{equation} \label{Recall}
\text{Recall} = \frac{\text{TP}}{\text{TP} + \text{FN}}
\end{equation}

\textbf{Average Precision (AP):} represents the area under the precision-recall curve, calculated using Equation \ref{AP}:
\begin{equation} \label{AP}
\text{AP}=\int_{0}^{1} \text{Precision}(\text{Recall}) \, d(\text{Recall})
\end{equation}

\textbf{Mean Average Precision (mAP):} represents the average AP value across all categories, indicating the model's overall detection performance across the entire dataset. This calculation is defined by Equation \ref{MAP}:
\begin{equation} \label{MAP}
\text{mAP} = \frac{1}{N} \sum_{i=1}^{N} \text{AP}_i
\end{equation}
where $\text{AP}_i$ represents the average precision value for the category indexed by $i$, and $N$ denotes the total number of categories in the training dataset.

\textbf{mAP$_{0.5}$:} is the average precision calculated at an IoU threshold of 0.5.

\textbf{mAP$_{0.5:0.95}$:} is calculated across IoU thresholds from 0.5 to 0.95, with values computed at intervals of 0.05.

\subsection{Experiment Results}
This section presents a comprehensive evaluation of the SOD-YOLOv8 model through targeted experiments. We begin by comparing the PIoU loss function with other common loss functions on YOLOv8s. Next, we assess the integration of the GFPN structure with the EMA and other attention modules. We then evaluate SOD-YOLOv8s against various YOLO variants (YOLOv3, YOLOv5s, YOLOv7) and widely used models (Faster R-CNN, CenterNet, Cascade R-CNN, SSD). Ablation studies validate the contributions of each enhancement. Visual experiments with the VisDrone2019 dataset demonstrate the model's effectiveness in diverse scenarios, including distant, high-density, and nighttime conditions. Finally, real-world traffic scene evaluations highlight the model's applicability and performance in challenging environments with cameras mounted on buildings at significant distances from the objects of interest.

\subsubsection{Comparative Evaluation of Bounding Box Regression}
\label{sssec:subsubhead}

To evaluate the impact of PIoU, we conducted comparative experiments on YOLOv8s using PIoU and other common loss functions under consistent training conditions. As shown in Table \ref{table:loss_comparison}, PIoU achieves the best detection performance. Specifically, it improves mAP$_{0.5}$ by 1.1\%, mAP$_{0.5:0.95}$ by 0.2\%, precision by 1.6\%, and recall by 0.4\% compared to CIoU. Additionally, the simpler loss function of PIoU makes model tuning easier, demonstrating its potential as an efficient and effective bounding box regression method.

\begin{table*}[t]
\centering
\caption{Detection results of YOLOv8s with different bounding box loss functions, shown as percentages (best outcomes in bold).}
\label{table:loss_comparison}
\begin{adjustbox}{max width=\textwidth} 
\begin{tabular}{@{}lcccc@{}}
\toprule
\textbf{Metrics} & \textbf{Precision\%} & \textbf{Recall\%} & \textbf{mAP$_{0.5}$\%} & \textbf{mAP$_{0.5:0.95}$\%} \\ \midrule
CIOU & 51.2 & 40.1 & 40.6 & 24 \\
EIOU & 49.8 & 39.9 & 40 & 23 \\
WIOU v1 & 51.4 & 40 & 40.7 & 23.7 \\
WIOU v2 & 51.9 & 40 & 40.7 & 23.8 \\
WIOU v3 & 52.6 & 40 & 41.2 & \textbf{24.2} \\
MPDIOU \cite{siliang2023mpdiou} & 52.1 & 39 & 40.7 & 23.9 \\
PIoU ($\lambda = 1.2$) & \textbf{52.8} & \textbf{40.5} & \textbf{41.7} & \textbf{24.2} \\ \bottomrule
\end{tabular}
\end{adjustbox}

\end{table*}

\begin{table*}[t] 
\centering
\caption{Detection results of GFPN-YOLOv8s with various attention mechanisms are presented as percentages. (Best results are highlighted in bold))
\label{table:attention_comparison}}
\begin{adjustbox}{max width=\textwidth}
\begin{tabular}{@{}lcccccccccccc@{}}
\toprule
\textbf{Models} & \textbf{Pedestrian} & \textbf{People} & \textbf{Bicycle} & \textbf{Car} & \textbf{Van} & \textbf{Truck} & \textbf{Bus} & \textbf{Motorcycle} & \textbf{mAP$_{0.5}$} \\ 
\midrule
YOLO-GFPN-SE & 50.9 & 42.6 & 15.0 & 83.5 & 45.1 & 39.2  & 53.4 & 51.3 & 42.9 \\
YOLO-GFPN-CBAM & 48.8 & 40.3 & 14.6 & 82.9 & 46.0 & 38.8  & 55.4 & 49.3 & 42.3 \\
YOLO-GFPN-CA & 51.0 & 42.3 & \textbf{17.8} & 83.8 & \textbf{47.5} & \textbf{40.1}  & 57.3 & 52.7 & 43.9 \\
\textbf{YOLO-GFPN-EMA} & \textbf{51.0} & \textbf{43.3} & 16.7 & \textbf{83.8} & 46.7 & 39.1  &  \textbf{60.9} & \textbf{52.8} & \textbf{44.2} \\ 
\bottomrule
\end{tabular}
\end{adjustbox}

\end{table*}

\subsubsection{Comparative experiment of attention mechanisms}

To evaluate the effectiveness of integrating the GFPN structure with the EMA attention mechanism, we incorporated three widely used attention modules—CBAM \cite{woo2018cbam}, CA \cite{hou2021coordinate}, and SE \cite{hu2018squeeze}—at the same position within the GFPN structure. This setup allowed for a direct comparison in our experiments, as detailed in Table \ref{table:attention_comparison}. The experimental results demonstrate that training with the GFPN-EMA combination consistently outperforms GFPN configurations with CBAM, CA, and SE in terms of mAP$_{0.5}$ values. Specifically, the GFPN-EMA model exhibits significant improvements across most object categories, particularly in Pedestrian, People, Car, Bus, and Motor classes, as well as overall mAP$_{0.5}$. As shown in Table \ref{table:attention_comparison}, these findings highlight EMA's efficacy in improving small object detection accuracy in the VisDrone dataset. EMA achieves this by addressing spatial interactions, overcoming 1x1 kernel convolution limitations through the integration of a 3x3 convolution for multi-scale spatial information, and employing cross-spatial learning to merge attention maps from parallel subnetworks. This approach effectively combines global and local spatial contexts.

\subsubsection{Comparison with different mainstream model}

Comparing SOD-YOLOv8s with other YOLO variants such as YOLOv3, YOLOv5s, and YOLOv7, SOD-YOLOv8s is remarkably efficient. Despite having a small model size of 11.5 million parameters, SOD-YOLOv8s achieves the highest accuracy metrics. It outperforms YOLOv3, YOLOv5s, and YOLOv7, despite YOLOv3 and YOLOv5s having larger model sizes ranging from 12.0 million to 18.3 million parameters. The YOLOv8 model has been adapted into various scales (YOLOv8n, YOLOv8s, YOLOv8m, YOLOv8l, and YOLOv8x) by adjusting width and depth, each progressively consuming more resources to improve detection performance. We conducted comparisons between SOD-YOLOv8s and different scales of YOLOv8 to further validate the performance of our proposed approach. Based on the information presented in  Table \ref{table:yolo-results}, Despite its lower parameter count of 11.5 million, SOD-YOLOv8s achieves highest recall (43.9\%), mAP$_{0.5}$ (45.1\%), and mAP$_{0.5:0.95}$ (26.6\%). In contrast, YOLOv8m, which has 25.9 million parameters, achieves lower accuracy metrics. This indicates that SOD-YOLOv8s is efficient in terms of computing capacity and model size, while also performing well in object detection tasks.

\begin{table*}[t] 
\centering
\caption{Different YOLO models' results, presented as percentages.(The best-performing outcomes are highlighted in bold)}
\label{table:yolo-results}
\begin{adjustbox}{max width=\textwidth}
\begin{tabular}{@{}lcccccccc@{}}
\toprule
Models & Model's Size & Backbone & Precision & Recall & mAP$_{0.5}$ & mAP$_{0.5:0.95}$ & Time/ms & Parameter/$10^6$ \\
\midrule
YOLOv3 \cite{redmon2018yolov3} & - & Darknet-53 & 53.6 & 43.2 & 42 & 23.1 & 209 & 18.3 \\
YOLOv5s & - & CSP-Darknet-53 & 46.7 & 34.8 & 34.7 & 19.2 & 13.9 & 12.0 \\
YOLOv7 \cite{wang2023yolov7} & - & ELAN & 51.5 & 42.3 & 40.1 & 21.8 & 71.5 & \textbf{1.7} \\
\multirow{3}{*}{YOLOv8 \cite{reis2023realtime}} & YOLOv8n & \multirow{3}{*}{CSP-Darknet-53} & 44.0 & 33.2 & 33.5 & 19.5 & \textbf{6.7} & 4.2 \\
                         & YOLOv8s &  & 51.1 & 39.1 & 39.6 & 23.8 & 7.8 & 11.1 \\
                         & YOLOv8m &  & \textbf{55.8} & 42.6 & 44.5 & \textbf{26.6} & 16.8  & 25.9 \\

SOD-YOLOv8s & - & CSP-Darknet-53 & 53.9 & \textbf{43.9} & \textbf{45.1} & \textbf{26.6} & 17.7 & 11.5 \\
\bottomrule
\end{tabular}
\end{adjustbox}
\end{table*}

This study conducted a comparative experiment to evaluate the performance of SOD-YOLOv8s against widely adopted models, including Faster R-CNN, CenterNet, Cascade R-CNN, and SSD. In Faster R-CNN \cite{ren2017faster}, the Region Proposal Network (RPN) \cite{ren2015faster} relies on backbone network features to generate region proposals. However, due to lower feature map resolution for small objects, the RPN may struggle to accurately localize them, leading to potential missed detections. Cascade R-CNN \cite{cai2018cascade} enhances detection performance through a multilevel architecture, albeit at the cost of increased computational complexity and training difficulty. CenterNet \cite{duan2019centernet} simplifies architecture with an anchor-free, center-point approach but faces challenges in precisely locating small objects in crowded or obscured scenes. These challenges arise from ambiguous object centers, interference from larger objects, complex backgrounds obscuring object centers, and sensitivity to pixel-level inaccuracies. Additionally, SSD's performance decreases on smaller objects compared to larger ones, as its shallow neural network layers may lack detailed high-level features necessary for accurate small object prediction. According to the data provided in Table \ref{table:models-results}, the SOD-YOLOv8s model achieves the highest performance in AP$_{0.5}$ of 45.1\% and AP$_{0.5:0.95}$ of 26.6\% compared to other models such as CenterNet, Cascade R-CNN, SSD, and Faster R-CNN.

\begin{table}[t] 
\centering
\caption{Results from different widely used models, presented as percentages.(The best-performing outcomes are highlighted in bold)}
\label{table:models-results}
\begin{tabular}{lcccc}
\toprule
Models                & Backbone & AP$_{0.5}$ & AP$_{0.5:0.95}$ \\
\midrule
Faster R-CNN \cite{ren2017faster}      &    ResNet       & 37.8          & 21.5               \\
Cascade R-CNN \cite{cai2018cascade}    &     ResNet     & 39.4          & 24.2               \\
CenterNet \cite{duan2019centernet}  &    ResNet50 \cite{he2016deep}     & 39.1        & 22.8               \\
SSD \cite{liu2016ssd}        &     MobileNetV2 \cite{sandler2018mobilenetv2}	     & 33.7          & 19               \\
SOD-YOLOv8s           &   CSP-Darknet-53  &   \textbf{45.1}          & \textbf{26.6}     \\
\bottomrule
\end{tabular}
\end{table}

\subsubsection{Ablation Experiments}

To validate the efficacy of each proposed enhancement approach in this study, ablation experiments were conducted on the baseline model. The results from these tests, shown in Table \ref{table: Ablation-classes}, demonstrate that each enhancement significantly improves detection performance across various categories. Introducing PIoU for bounding box regression enhances localization without enlargement issues, leading to a significant 1.1\% increase in $mAP_{0.5}$. This improvement is particularly beneficial for categories such as Pedestrian, People, and Bicycle. Integrating an enhanced GFPN and incorporating a new small object detection layer into the YOLOv8 network results in a significant 2.9\% increase in $mAP_{0.5}$. This enhancement demonstrates substantial performance improvements across all categories, including Pedestrian, Bicycle, Car, Van, and Motor, highlighting GFPN's effectiveness in capturing multi-scale features. Furthermore, integrating the C2f-EMA module, which utilizes the EMA attention mechanism, and replacing C2f with it within the neck layers increases $mAP_{0.5}$ by 0.5\%. This enhancement notably benefits categories such as People, Motor, and Truck, demonstrating its effectiveness in improving detection across various categories. According to Table \ref{table:comparison}, our proposed efficient model enhances object detection performance significantly without adding significant computational cost or latency compared to YOLOv8s. It improves recall from 43\% to 44\%, precision from 45\% to 46\%, $mAP_{0.5}$ from 40\% to 45.1\%, and $mAP_{0.5:0.95}$ from 20\% to 26.6\%.

\begin{table*}[t] 
\centering
\caption{Comparative experiments between the enhanced model and YOLOv8s across various categories, with percentages presented (best-performing outcomes highlighted in bold).\label{table: Ablation-classes}}
\begin{adjustbox}{max width=\textwidth}
\begin{tabular}{@{}lccccccccccccc}
\toprule
\textbf{Models} & \textbf{Pedestrian} & \textbf{People} & \textbf{Bicycle} & \textbf{Car} & \textbf{Van} & \textbf{Truck} & \textbf{Bus} & \textbf{Motorcycle} & \textbf{mAP$_{0.5}$}  \\ 
\midrule
YOLOv8s & 43.5 & 34.2 & 14.9 & 79.5 & 45.0 & 40.3  & 58.1 & 45.4 & 40.6  \\ 
YOLOv8s-PIoU & 46 & 38.2 & 15.6 & 80.4 & 45.8 & 39 & 60.2 & 47 & 41.7 \\ 
YOLOv8s-PIoU-GFPN & 52.8 & 44.0 & 17.7 & \textbf{84.2} & \textbf{47.7} & 39.4 & 60.8 & 53.2 & 44.6  \\ 
YOLOv8s-PIoU-GFPN-EMA & \textbf{53.1} & \textbf{44.5} & \textbf{18.2} & 83.9 & 47.1 & \textbf{41.0}  & \textbf{60.9} & \textbf{53.8} & \textbf{45.1}  \\ 

\bottomrule

\end{tabular}
\end{adjustbox}
\end{table*}

\begin{table*}[t] 
\centering
\caption{ Detection results following the adoption of different improvement strategies, presented as percentages.(The best-performing outcomes are highlighted in bold)\label{table:comparison}}
\begin{adjustbox}{max width=\textwidth}
\begin{tabular}{@{}lccccccccccc@{}}
\toprule
\textbf{Baseline} & \textbf{PIoU} & \textbf{GFPN} & \textbf{EMA}  & \textbf{Precision} & \textbf{Recall} & \textbf{mAP$_{0.5}$} & \textbf{mAP$_{0.5:0.95}$} &  \textbf{Detection Time/ms} & \textbf{Parameter/${10}^{6}$} \\
\midrule
\checkmark &  &  &    & 51.2 & 40.1 & 40.6 & 24 & 7.8 & 11.1 \\
\checkmark & \checkmark &  &   & 52.8 & 40.5 & 41.7 & 24.2  & 7.4 & 11.1 \\
\checkmark & \checkmark & \checkmark &    & 52.7 & 44.3 & 44.6 & 26.3  & 11.5 & 11.5 \\
\checkmark & \checkmark & \checkmark & \checkmark   & 53.9 & 43.9 & 45.1 & 26.6  & 11.6 & 11.5 \\

\bottomrule
\end{tabular}
\end{adjustbox}
\end{table*}

\begin{figure}
    \centering
    \includegraphics[width=1\linewidth]{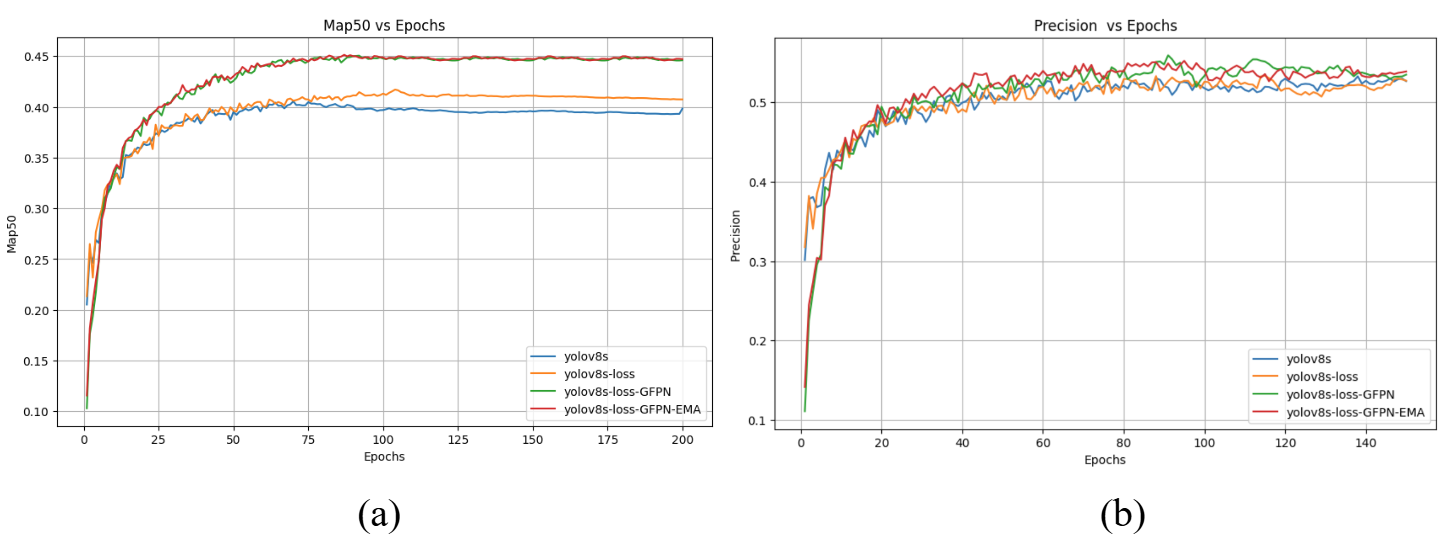}
    \caption{(a) Training progress plot comparing YOLOv8s-GFPN-EMA, YOLOv8s-GFPN, and YOLOv8s based on $mAP_{0.5}$ (b) and precision}
    \label{fig:Training-progress}
\end{figure}

Figure \ref{fig:Training-progress} depicts the evaluation metrics for SOD-YOLOv8 and YOLOv8s across 200 training epochs. Our model outperforms YOLOv8s in precision and $mAP_{0.5}$ starting around epoch 15 and stabilizes after 50 epochs. This illustrates that SOD-YOLOv8 significantly enhances detection performance, particularly for small and challenging objects, without introducing significant complexity.

\subsubsection{Visual assessment}

We conducted visual experiments to evaluate our model's detection performance. Our analysis included various metrics such as confusion matrices and inference test results. To validate the effectiveness of our method in challenging real-world scenarios, we performed inference tests using images captured by a camera mounted on the 12th floor of a building. This scenario involves capturing images from a high vantage point, posing challenges in detecting numerous small objects in a crowded traffic scene at an intersection.\\

\textbf{VisDrone2019 dataset results}\\

To visualize the performance of SOD-YOLOv8s, we utilize a confusion matrix. This matrix organizes predictions into a format where each row corresponds to instances of a true class label, and each column corresponds to instances predicted by the model. Diagonal elements indicate correct predictions, where the predicted class matches the actual class. Off-diagonal elements represent incorrect predictions, where the predicted class does not match the actual class.

\begin{figure}
    \centering
    \includegraphics[width=1\linewidth]{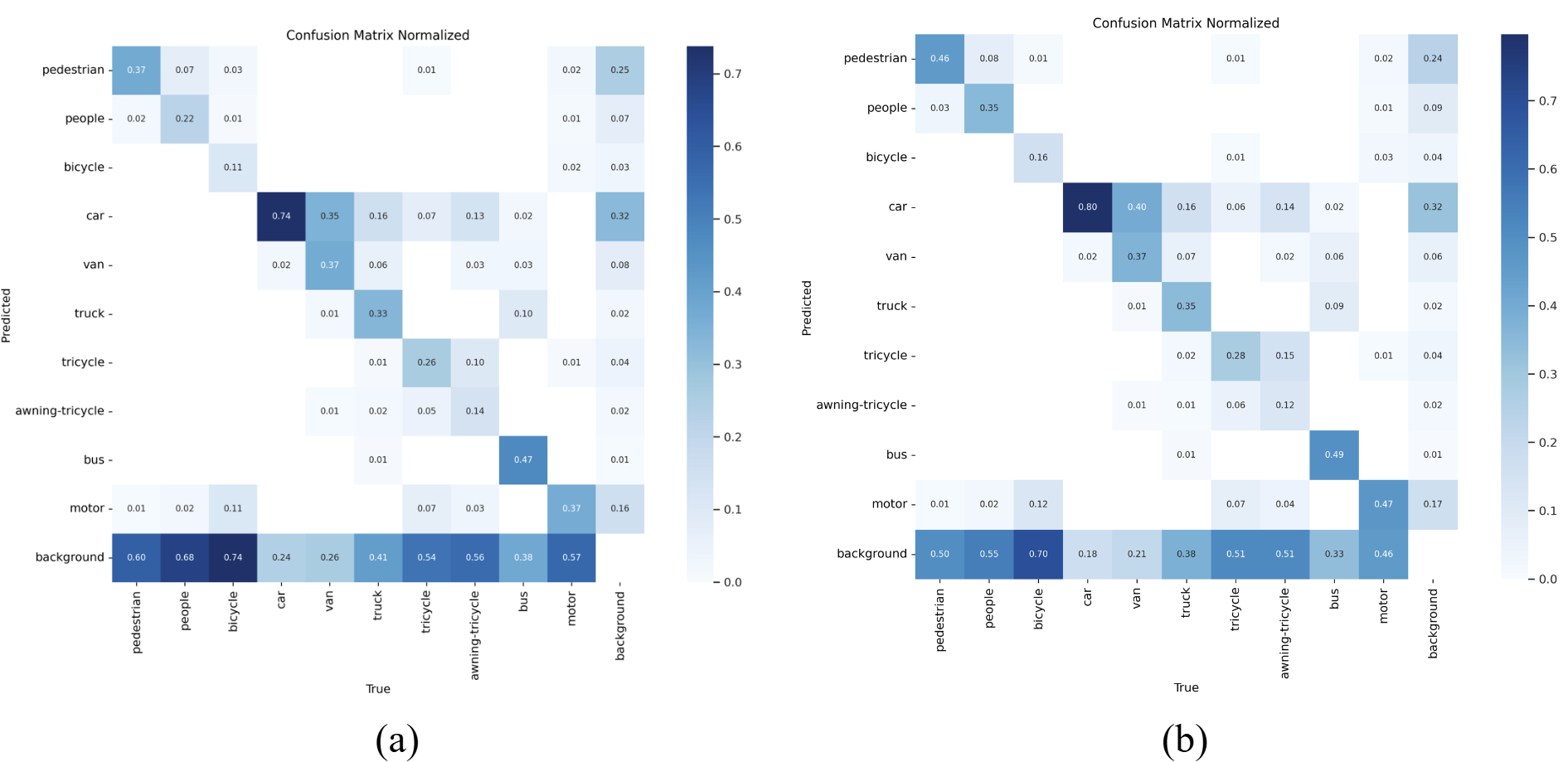}
    \caption{Confusion matrix of YOLOv8s; (b) confusion matrix of proposed model.}
    \label{fig:confusionMatrix_compare}
\end{figure}

Figure \ref{fig:confusionMatrix_compare} demonstrates improved detection performance of SOD-YOLOv8s across most object categories. The confusion matrix shows lighter shades in the last row compared to YOLOv8s, indicating reduced misclassifications of objects as background. However, challenges remain in accurately identifying bicycles, tricycles, and awning-tricycles, which are often mislabeled as background. Despite these issues, SOD-YOLOv8s shows darker shades along the main diagonal, indicating an overall increase in correctly detected objects.

\begin{figure}
    \centering
    \includegraphics[width=1\linewidth]{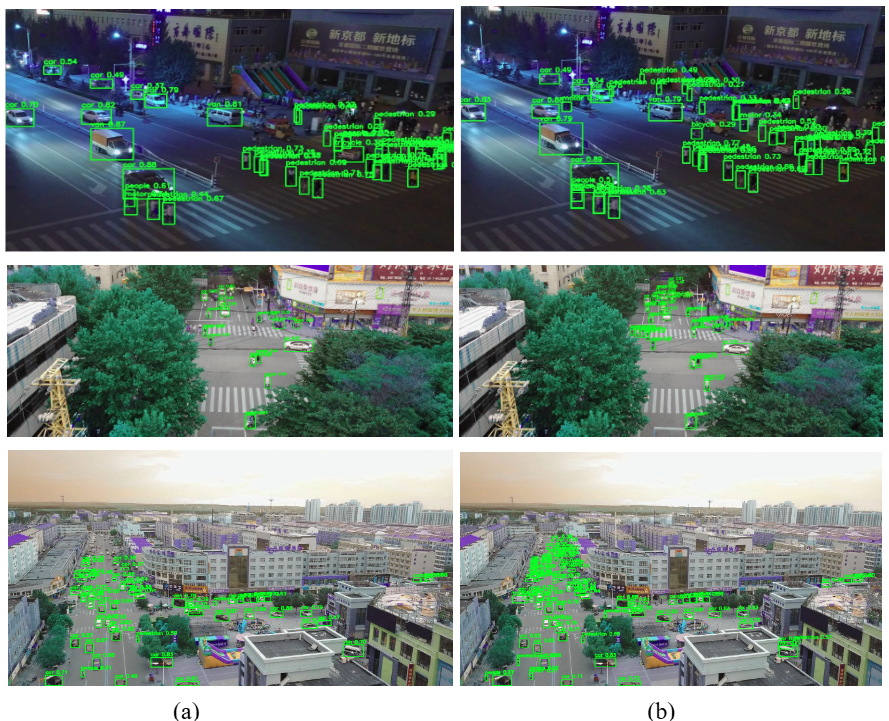}
    \caption{Inference results for (a) YOLOv8s and (b) SOD-YOLOv8s across diverse scenarios including distant and high-density objects, as well as nighttime scenarios, using the VisDrone2019 dataset.}
    \label{fig:Inference-VisDrone}
\end{figure}

As depicted in Figure \ref{fig:Inference-VisDrone}, we assess the efficacy of SOD-YOLOv8 across three challenging scenarios within the Visdrone dataset: nighttime conditions, crowded scenes with high-density objects, and scenes with distant objects. Remarkably, across all three scenarios, notable improvements are observed. In the nighttime scenario, illustrated in the first row of Figure \ref{fig:Inference-VisDrone}, objects are detected with higher IoU values, and a greater number of smaller objects are successfully identified. In the second scenario, depicted in the second row of Figure \ref{fig:Inference-VisDrone}, SOD-YOLOv8 demonstrates superior performance by successfully detecting numerous small objects located at the corners of intersections, a task which YOLOv8s struggles with. Similarly, in the third scenario involving objects positioned farther from the camera, SOD-YOLOv8s excels in detecting objects with higher IoU values and successfully identifying a greater number of smaller objects. These results demonstrating the substantial improvements provided by SOD-YOLOv8s across different environmental conditions, indicating its reliability and effectiveness in detecting objects in challenging scenarios.\\

\textbf{Real dataset results}\\

\begin{figure}
    \centering
    \includegraphics[width=1\linewidth]{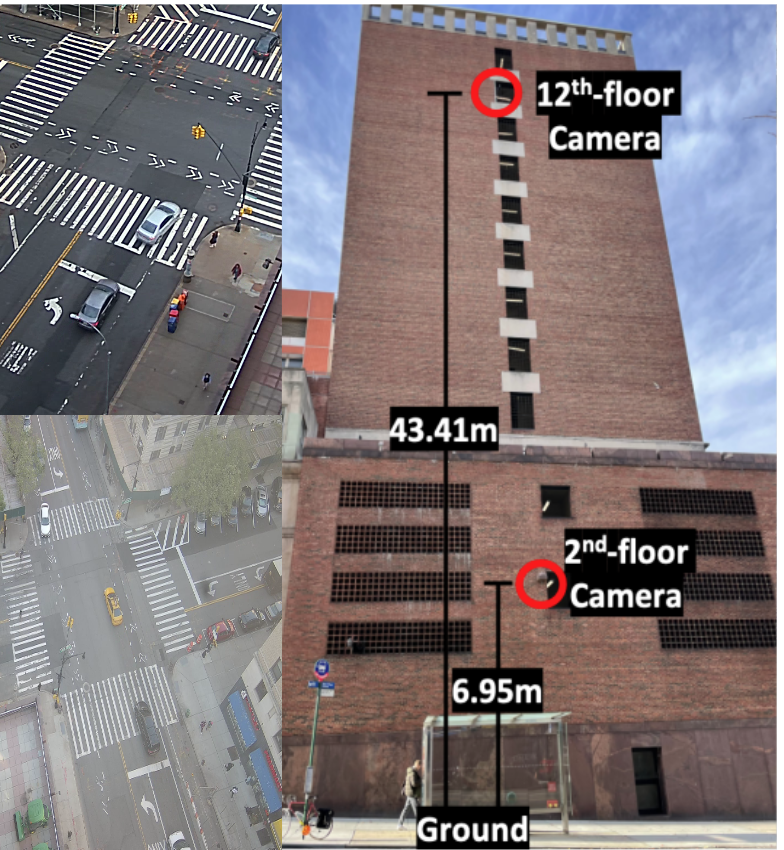}
    \caption{The perspective captured by COSMOS cameras on the 12th floor of Columbia’s Mudd building overlooking the intersection \cite{cosmos2022}.}
    \label{fig:cameras_deployed}
\end{figure}

This section evaluates the model's performance in dynamic, real-world challenging scenarios where the camera is mounted on a building at a significant distance from the objects of interest. To assess the applicability and generalization of the proposed SOD-YOLOv8 model, we conducted inference experiments using real-world data from a traffic scene scenario. The image data were primarily captured by NSF PAWR COSMOS testbed cameras \cite{cosmos2020, kostic2022, cosmos2022} mounted on the 12th floor of Columbia’s Mudd building (Fig. \ref{fig:cameras_deployed}), overlooking the intersection of 120th St. and Amsterdam Ave. in New York City. Images were specifically selected from this vantage point to utilize its elevated perspective and greater distance from the street. This viewpoint poses a unique challenge for object detection, requiring enhanced perception due to the reduced scale of objects, including various vehicle types and pedestrians.

\begin{figure}
    \centering
    \includegraphics[width=1\linewidth]{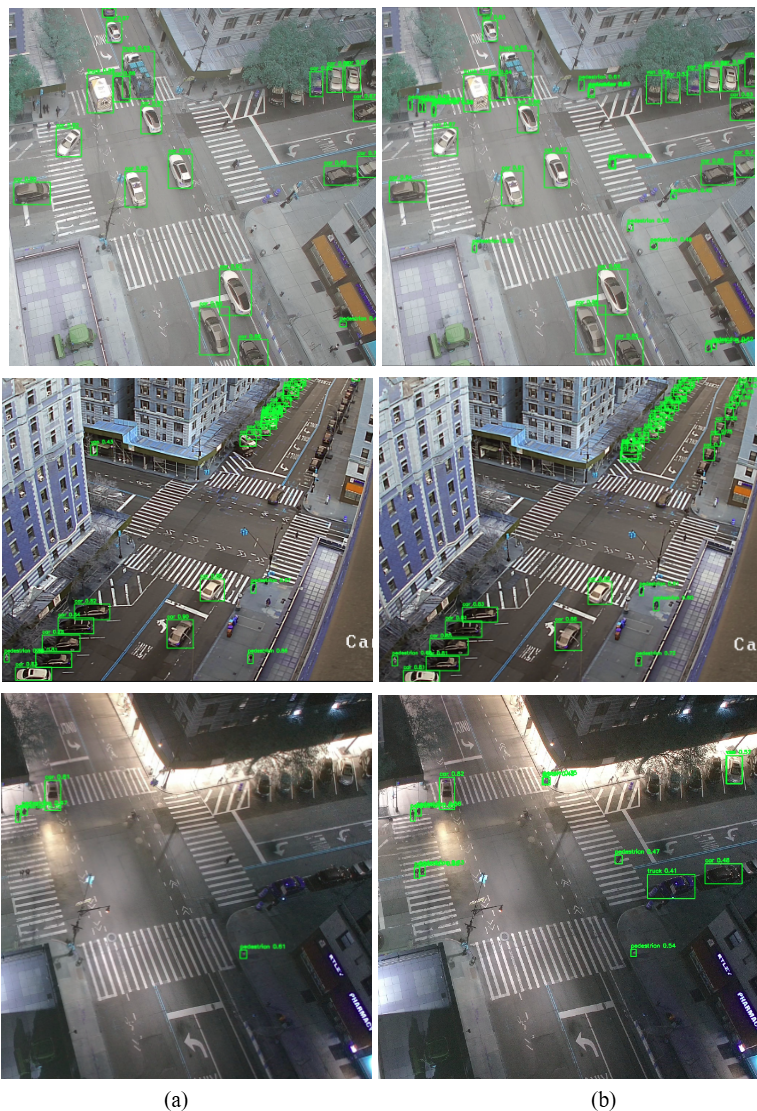}
    \caption{Inference results for (a) YOLOv8s and (b) SOD-YOLOv8s across various scenarios, including scenes with distant and high-density objects, as well as nighttime scenarios, using the traffic scene dataset.}
    \label{fig:real_inference}
\end{figure}

As depicted in Figure \ref{fig:real_inference}, we assess SOD-YOLOv8's performance in three challenging real-world traffic scenarios using images from cameras on the 12th floor such as crowded scenes with high-density objects, distant objects, and nighttime conditions. Significant improvements are observed across all three scenarios. In the first scenario, shown in the top row of Figure \ref{fig:real_inference}, SOD-YOLOv8 outperforms YOLOv8s by successfully detecting numerous small-scale pedestrians at the corners of intersections, a task where YOLOv8s struggles. In the second scenario, with distant objects, SOD-YOLOv8 shows superior performance, achieving higher IoU values and effectively detecting more small objects. In the nighttime scenario, shown in the third row of Figure \ref{fig:real_inference}, SOD-YOLOv8 achieves higher IoU values for detected objects and identifies more small objects compared to the YOLOv8s baseline model, despite challenging lighting conditions. These results illustrate the substantial improvements achieved by SOD-YOLOv8 across diverse environmental conditions, highlighting its reliability and effective object detection capabilities in challenging scenarios. \\

\section{Conclusion}

Detecting small-scale objects in traffic scenarios presents significant challenges that can reduce overall effectiveness. To address these issues, we introduced SOD-YOLOv8, a specialized object detection model designed for aerial photography and traffic scenes dominated by small objects. Built upon YOLOv8, this model integrates enhanced multi-path fusion inspired by the GFPN architecture of DAMO-YOLO models, facilitating effective feature fusion across layers and simplifying architecture through reparameterization. By leveraging a high-resolution fourth layer and incorporating a C2f-EMA structure, SOD-YOLOv8 prioritizes small objects, enhances feature fusion, and improves precise localization. Also PIoU is used as a replacement for CIoU, the IoU-based loss function in YOLOv8.

The SOD-YOLOv8 model outperforms widely used models such as CenterNet, Cascade R-CNN, SSD, and Faster R-CNN across various evaluation metrics. Our efficient model significantly enhances object detection performance without substantially increasing computational cost or detection time compared to YOLOv8s. It improves recall from 40.1\% to 43.9\%, precision from 51.2\% to 53.9\%, $mAP_{0.5}$ from 40.6\% to 45.1\%, and $mAP_{0.5:0.95}$ from 24\% to 26.6\%. In real-world traffic scenarios captured by building-mounted cameras, SOD-YOLOv8 achieves higher IoU values and identifies more small objects than YOLOv8s, even under challenging conditions like poor lighting and crowded backgrounds. These capabilities make it ideal for applications such as UAV-based traffic monitoring.

However, challenges remain in deploying small object detection methods in resource-constrained environments. While attention mechanisms and complex feature fusion improve performance in controlled settings, they may struggle with generalization across diverse environments and conditions, complicating real-world deployment and maintenance. In this study, given the promising results of the used PIoU method on the VisDrone dataset and real-world traffic scenes, which involve numerous small objects, future research will prioritize evaluating PIoU across various datasets. Additionally, efforts will focus on refining the GFPN architecture, exploring alternative processing methods, and assessing the model's performance in adverse weather conditions to enhance its adaptability and robustness across diverse scenarios.

\section*{ACKNOWLEDGMENT}

This work was supported by the Center for Smart Streetscapes, an NSF Engineering Research Center, under grant agreement EEC-2133516. The authors are grateful to Eric Valasek and Nicholas D'Andre from Gridmatrix for motivating the small object detection problem through a number of discussions

\end{document}